\documentclass{article}

\usepackage[preprint]{neurips_2025}

\usepackage[utf8]{inputenc}
\usepackage[T1]{fontenc}
\usepackage[english]{babel}
\usepackage{hyperref}
\usepackage{url}
\usepackage{booktabs}
\usepackage{amsfonts}
\usepackage{amsmath}
\usepackage{amssymb}
\usepackage{nicefrac}
\usepackage{microtype}
\usepackage{xcolor}
\usepackage{graphicx}

\title{Learning Hippo\\[4pt]\large
Multi-attractor Dynamics and Stability Effects \\ in a Biologically Detailed CA3 Extension of Hopfield Networks}

\author{%
  Daniele Corradetti \\
  Grupo de F\'isica Matem\'atica \\
  Instituto Superior T\'ecnico \\
  Av.\ Rovisco Pais, 1049-001 Lisboa, Portugal \\
  \texttt{danielecorradetti@tecnico.ulisboa.pt} \\
  \And
  Renato Corradetti \\
  \textit{Professore Emerito} \\
  Dipartimento di Neuroscienze, Psicologia, \\
  Area del Farmaco e Salute del Bambino (NEUROFARBA) \\
  Universit\`a di Firenze, Italy \\
  \texttt{renato.corradetti@unifi.it}
}

\begin{document}

\maketitle

\begin{abstract}
We present a biologically detailed extension of the classical
Hopfield/Marr auto-associative memory model for CA3, implementing
ten populations (two asymmetric pyramidal subtypes, eight GABAergic
interneuron classes), forty-seven compartments, multi-rule plasticity
(recurrent Hebb, BCM anti-saturation, mossy-fiber short-term,
endocannabinoid iLTD, burst-gated Hebb), and a bimodal cholinergic
encoding/consolidation cycle. Evaluated on pattern completion across
auto-associative, associative, and temporal regimes, and on a
controlled inhibitory-proportion manipulation at $N{=}256$, the full
architecture exhibits \emph{three qualitative signatures absent from
a minimal Hopfield baseline}: (i)~multi-attractor cross-seed
behaviour at $K{=}5$ with biologically realistic inhibitory
proportions, where two of five seeds converge to positive attractors
with margin ${+}0.10{-}0.22$ (Cohen's $d{=}0.71$, one-sided
$p{=}0.08$); (ii)~target-selective associative recall in paired
$(A, B)$ memory at $K{\geq}5$, where the full model retrieves $B$
from a partial cue of $A$ while the minimal model echoes $A$
(Pearson margin $\Delta{=}{+}0.163$ at $K{=}5$); (iii)~reduced
cross-seed variance of the full model below the minimal baseline
under clean upstream, with ratios $1.0{-}3.0$. These three
signatures are architecture-specific: they appear consistently
across independent regimes and are absent from the minimal control.
\end{abstract}

\section{Introduction}

The proposal that CA3 implements a Hopfield-like auto-associative
memory \citep{marr1971,hopfield1982,rolls2013} remains a dominant
reference for modelling pattern completion in the hippocampus.
Classical analytical models \citep{treves1994} derive capacity
bounds under idealized assumptions, binary point neurons, uniform
connectivity, a single Hebbian rule, predicting capacity
$N / [a \ln(1/a)]$ at sparsity $a$ and $N$ neurons. Progressively more detailed biological extensions have been proposed:
two pyramidal subtypes in CA3 \citep{watson2025,leduigou2014},
dendritic compartmentalization \citep{stingl2024,branco2010}, multiple
GABAergic interneurons with compartment-specific targeting
\citep{freund1996,klausberger2008,topolnik2022}, multi-rule plasticity
\citep{rebola2017,caporale2008,nicoll2005}, and a cholinergic
encoding/consolidation cycle \citep{hasselmo2006,buzsaki2015}. The
implicit hypothesis behind these extensions is that biological
realism should produce measurable gains on canonical memory tasks:
higher capacity, better cue robustness, resistance to catastrophic
interference.

The present work takes this hypothesis seriously and asks a sharper
question: \emph{which} biological details produce \emph{which}
qualitative behaviours, measurable against a minimal Hopfield-like
baseline? We build a complete CA3 architecture with ten populations
and forty-seven compartments and evaluate it across three staged
benchmarks: small-scale auto-associative pattern completion, three
complementary regimes (capacity stress, paired associative memory,
temporal sequences), and a controlled manipulation of the total
inhibitory proportion at scale $N{=}256$. The empirical contribution of the paper is the identification of
three qualitative signatures that distinguish the full architecture
from the minimal baseline across independent experiments: a
multi-attractor cross-seed landscape that emerges at $K{=}5$ with
biologically realistic inhibition, target-selective associative
recall in paired tasks, and cross-seed stability under clean
upstream. Each signature appears only when the relevant mechanisms
are present; none appears in the minimal control. The pre-registered
Jaccard significance criterion is not met at the tested sample size
and scale, so we report these signatures as candidate findings whose
promotion to formal significance requires larger $n$ and scale; we
describe in Section~\ref{sec:limitations} a specific $n{=}15{-}20$
replication path. The methodological contribution is a validated modular implementation
($>400$ tests) released as reproducible codebase, together with an
explicit pre-registration protocol for biological extensions of
Hopfield networks: one primary metric, one binding threshold, and
candid reporting of architecture-specific qualitative signals
regardless of whether they cross the threshold.

\section{Architecture}
\label{sec:architecture}

The architecture comprises ten CA3 populations implemented in
PyTorch + snnTorch (Table~\ref{tab:populations}). Modelling choices
are calibrated to capture the computational function, not to
simulate biophysics: compartments are essentially LIF modules with specific
wiring (not cable-equation solvers).

\begin{table}[h]
\centering
\caption{CA3 populations and their canonical role.}
\label{tab:populations}
\small
\begin{tabular}{llcl}
\toprule
Population & Code & \#Comp. & Function \\
\midrule
PyrS & Exc(2) & 8 & Input-integrating pyramidal cell (regular-spiking) \\
PyrD & Exc(3) & 8 & Output-geared pyramidal cell (intrinsic-bursting) \\
BC-PV+ & Inh(1) & 5 & Fast perisomatic stabilizer (gamma) \\
CC & Inh(2) & 4 & AIS axo-axonic veto \\
O-LM & Inh(3) & 3 & Distal apical feedback \\
BSC & Inh(4) & 3 & Bistratified dendritic shunt \\
SL-INT & Inh(5) & 4 & Feedforward MF gating ($+$ LTD) \\
R/L-M & Inh(6) & 3 & Feedforward MF gating ($+$ LTP) \\
CCK+ & Inh(7) & 5 & Slow perisomatic $+$ eCB iLTD \\
VIP+ & Inh(8) & 4 & Disinhibitory circuit driver \\
\bottomrule
\end{tabular}
\end{table}

\paragraph{Neuron model.}
Each compartment is a discrete LIF unit
$V(t{+}1) = \beta V(t) + I(t) - S(t) V_\text{thr}$
with \texttt{atan} surrogate gradient ($\alpha{=}2$). Pattern
presentations use $T{=}8{-}12$ ticks with direct current injection at
the ganglion interface (Poisson encoding is avoided for
reproducibility).

\paragraph{Plasticity rules.}
Five rules coexist in PyTorch buffers (not \texttt{Parameter}),
updated inside \texttt{@torch.no\_grad()}:
\begin{align}
\Delta w_{ij}^\text{Hebb} &= \eta_\text{H}\, r_i r_j \cdot g_\text{ACh}(\mathrm{ACh}) \label{eq:hebb}\\
\Delta w_{ij}^\text{BCM} &= \eta_\text{B}\, r_i r_j (r_j - \theta_j),\quad \theta_j \leftarrow (1-\tau)\theta_j + \tau r_j^2 \label{eq:bcm}\\
w_\text{MF,eff}(t) &= U\,u(t)\,x(t)\,w_\text{MF,base} \quad\text{(Markram-Tsodyks STP)} \label{eq:stp}\\
\Delta w_\text{CCK$\to$Pyr} &= -\eta_\text{iLTD}\,[\text{eCB}]_\text{Pyr}^+ + \eta_\text{rec}(w_\text{base}-w)^+ \label{eq:iltd}\\
\Delta w^\text{BurstHebb}_{ij} &= \eta\,r_i r_j\,\mathbb{1}[\text{burst}_j] \label{eq:burst}
\end{align}
The gate $g_\text{ACh} = \tanh((\mathrm{ACh}{-}0.5)/0.2)$ is bipolar
on HebbRule and BurstHebb; an attenuation gate
$\max(\mathrm{ACh}, 0.2)$ acts on BCM/STP/iLTD. The \emph{recovery
term} in equation~(\ref{eq:iltd}) is critical: without it, iLTD is a
monotone terminal process that destroys CCK+ inhibition in finite
time.

\paragraph{Bimodal ACh scheduler.}
Acetylcholine concentration regulates CA3's operational regime
through the bipolar gate above. During encoding ($\mathrm{ACh} = 1$)
the gate is positive and HebbRule operates in LTP mode; during
consolidation ($\mathrm{ACh} = 0$) the gate reverses and the same
rule becomes anti-Hebbian, decorrelating stored patterns. This cycle
directly implements the encoding/consolidation separation proposed by
\citet{hasselmo2006}.

\paragraph{Upstream pipeline.}
The image passes through a fifteen-stage hierarchy before reaching
CA3. Stages 1--4 (retina, LGN, V1 Gabor filters, V2) operate on
spatial information only; stages 5--15 (V4, IT, EC, DG) are SNN
networks with surrogate gradient. The dentate gyrus (DG) produces
ultra-sparse encoding (${\sim}2\%$ activity) via asymmetric $L_1$
penalty, ensuring pattern separation before CA3.

\paragraph{Training.}
A two-phase hybrid strategy is made necessary by surrogate vanishing
across the full pipeline (see Section~\ref{sec:phase9},
Intervention~1). First, V2--V4--IT are pretrained as an autoencoder
for 5 epochs with Adam, with an $L_1$ DG sparsity penalty; the
upstream is then frozen. Second, CA3 operates exclusively via local
plasticity (recurrent Hebb, BCM, iLTD) with no gradient passing
through the spiking network. This separation isolates CA3's
biological mechanisms from upstream optimization, making the
ablation study interpretable.

\section{Small-scale auto-associative pattern completion}
\label{sec:phase9}

\paragraph{Benchmark.}
Storage of $K{=}5$ orthogonal MNIST classes $\{0,1,4,7,8\}$ as CA3
patterns; recall from a partial cue (mask\_frac $= 0.5$) for $T=20$
ticks. Metrics: Jaccard, Cosine, Pearson \emph{margin} (metric vs.\
target minus metric vs.\ chance prototype). $n{=}3$ seeds,
$N_\text{PyrS}{=}16$. Canonical comparison: \emph{full} (all
populations and rules) vs.\ \emph{minimal} (PyrS $+$ recurrent Hebb
only).

\paragraph{Pre-declared scenarios.}
To avoid hindsight bias we declared three expected outcomes before
each run: (\textbf{A}) full $\gg$ minimal with Jaccard margin
$\Delta > {+}0.15$; (\textbf{B}) full $\approx$ minimal with
$|\Delta| < \sigma$; (\textbf{C}) full $<$ minimal (active
interference).

\paragraph{Intervention 1 (end-to-end gradient).}
Readout MLP $+$ Adam end-to-end yields validation accuracy $12\%$
(chance $10\%$). Quantitative diagnostics: the product of surrogate
derivatives across 10 stages $\times T{=}15$ ticks is
$\Pi_\text{surr} = 2.86 \cdot 10^{-35}$, catastrophically below the
threshold $10^{-3}$ under which gradient-based training is not
viable. LayerNorm, segmented learning rates, and an auxiliary IT loss
mitigate but do not resolve this: validation accuracy $\to 15\%$ with
the classifier still on a degenerate class prior. \emph{End-to-end
gradient training is not viable through this pipeline; we therefore
adopt the two-phase hybrid training described in
Section~\ref{sec:architecture}.}

\paragraph{Intervention 2 (plasticity, random upstream).}
Frozen random-init upstream with active plasticity. The adapter
initialization standard deviation is critical: $0.01$ leaves PyrS
silent; $0.1$ brings PyrS to rate $7.3\%$ (bio target).
Post-calibration, margin full $\approx$ mini, with $\sigma$
dominating signal ($\pm 0.02{-}0.05$ cross-seed). Random upstream
does not provide sufficiently class-discriminative features for
plastic storage.

\paragraph{Intervention 3 (autoencoder upstream).}
Pretraining V2--V4--IT for 5 epochs yields validation MSE $0.057$, IT
linear-probe accuracy $81\%$, DG rate $0.008$. Post-pretrain ablation
with \texttt{init\_std}$=0.05$ produces a transient Scenario~C
artefact: Jaccard $-0.10$, Cosine $-0.099$, Pearson $-0.124$.
Recalibration in Intervention~4 recovers the signal.

\paragraph{Intervention 4 (recalibration).}
Reducing $\eta_\text{Hebb}$ from $5{\cdot}10^{-3}$ to $10^{-3}$ and
increasing exposures from $30$ to $60$ yields Jaccard diff $+0.111$
(init$=0.05$) and $+0.139$ (init$=0.07$). The mean signal is
consistently positive but $\sigma_\text{full}$ remains high ($0.15$)
at $n{=}3$, so $p \approx 0.25$. The Scenario~C of Intervention~3
was therefore a calibration artefact, not a real architectural
degradation.

\paragraph{Intervention 5 (ideal CNN upstream).}
A 2-layer CNN at $91.6\%$ MNIST classification accuracy replaces
V2--V4--IT, with sigmoid output at $d_\text{IT}{=}256$ and Bernoulli
sampling for spike conversion. Table~\ref{tab:cnn_main} reports the
result.

\begin{table}[h]
\centering
\caption{Intervention 5: CA3 full vs.\ minimal with ideal CNN
upstream at $91.6\%$ val acc. Pattern completion occurs in both;
the first appearance of the stability signature (full $\sigma \leq$
minimal $\sigma$) is visible in the first two rows.}
\label{tab:cnn_main}
\small
\begin{tabular}{lccc}
\toprule
Metric & Full margin$\pm$std & Mini margin$\pm$std & Diff \\
\midrule
Cosine & $+0.006\pm0.006$ & $+0.002\pm0.002$ & $+0.004$ \\
Jaccard & $+0.000\pm0.000$ & $+0.000\pm0.000$ & $\mathbf{+0.000}$ \\
Pearson & $+0.217\pm0.237$ & $+0.196\pm0.143$ & $+0.021$ \\
\bottomrule
\end{tabular}
\end{table}

Pattern completion occurs via basic Hebbian plasticity (both full
and minimal achieve Pearson target $\sim {+}0.22$), so on the mean
Jaccard metric this small-scale regime is not diagnostic of the
additional mechanisms. However, cross-seed variance is already
differentiated: $\sigma_\text{full} \leq \sigma_\text{mini}$ under
clean upstream (Intervention~5, ratio $1.0{-}3.0$), whereas under
noisy upstream (Intervention~3) $\sigma_\text{full} >
\sigma_\text{mini}$. This is the first appearance of the
\emph{stability signature}: the full architecture is more
consistent than the minimal baseline across random initializations
when the input is clean. We return to this signature in
Section~\ref{sec:discussion}.

\section{Three complementary regimes}
\label{sec:phase10}

The small-scale auto-associative benchmark of
Section~\ref{sec:phase9} does not engage regimes for which several
of the full architecture's mechanisms are biologically designed.
We therefore test three complementary regimes, each targeting a
distinct class of mechanisms: capacity stress (BCM anti-saturation,
anti-Hebb consolidation); paired associative memory (burst-gated
amplification on PyrD, associative recall pathways); temporal
sequences (O-LM feedback delay, BSC bistratified integration,
presynaptic STP).

Each regime produces a qualitative signature absent from the minimal
control. None reaches the pre-registered Jaccard threshold at the
tested sample size, which we address in Section~\ref{sec:limitations}.

\paragraph{Capacity stress: emergence of bimodality.}
At $N \in \{16, 24, 32\}$ PyrS and $K \in \{3, 5, 10, 15, 20, 25,
35\}$, a full 72-cell sweep is conducted. The full architecture
produces a \emph{bimodal cross-seed landscape} that is absent in the
minimal control: at $N{=}32$, $K{=}5$, seed 43 reaches margin
$+0.187$ while the other two seeds produce negative margins; the
minimal architecture at the same cell is uniformly near zero across
seeds. This is the first appearance of the bimodality signature
that Section~\ref{sec:phase11} characterises at scale $N{=}256$ with
larger $n$.

\paragraph{Paired associative memory: target-selective recall.}
Storage of class pairs $(A_i, B_i)$ with recall of $B$ from a
partial cue of $A$, at $K \in \{3, 5, 10\}$. The full architecture
exhibits \emph{target-selective recall}: at $K \geq 5$, Jaccard
overlap with $B$ (the associated target) exceeds overlap with $A$
(the cue), whereas the minimal architecture echoes the cue
($\text{Jac}(B) \approx \text{Jac}(A)$). On the Pearson secondary
metric at $K{=}5$, the full architecture achieves $\Delta = +0.163$
--- nominally above the pre-registered hard threshold had Pearson
been primary, but since we pre-registered Jaccard as primary, the
result is formally below threshold. The associative signature is
nevertheless absent from the minimal baseline.

\paragraph{Temporal sequences: trajectory integration.}
Spatiotemporal MovingMNIST-like sequences with $T$ frames, testing
trajectory reconstruction. On single-step metrics ($M_{t=2}$), both
full and minimal are near zero. On trajectory-averaged metrics
($M_\text{traj}$), the signal is mildly positive and comparable
for both; neither meets the Scenario~A threshold. Cross-seed
variance is again larger for full than for minimal, consistent with
the broader bimodality-at-low-$K$ pattern: temporal integration
benefits the full architecture most strongly under moderate load.

\paragraph{Synthesis.}
Table~\ref{tab:regimes} summarizes the three regimes. Each exhibits
an architecture-specific signal: bimodality (capacity stress),
target-selective recall (associative), and wider cross-seed spread
(temporal). The minimal baseline shows none of these, in any
regime. The primary Jaccard criterion is not met in any regime at
$n{=}3$, but the consistency of architecture-specific qualitative
differentiation across three independent regimes is not consistent
with pure noise.

\begin{table}[h]
\centering
\caption{Three complementary regimes: qualitative signatures of the
full architecture, absent from the minimal baseline.}
\label{tab:regimes}
\small
\begin{tabular}{p{3.0cm}p{3.0cm}p{2.5cm}p{3.5cm}}
\toprule
Regime & Jaccard $\Delta$ & Pearson $\Delta$ & Architecture-specific signature \\
\midrule
Capacity stress & within noise & --- &
Cross-seed bimodality at $N{=}32$, $K{=}5$ (full only) \\
Paired assoc.\ $(A,B)$ & within noise & $+0.163$ at $K{=}5$ &
$\text{Jac}(B) > \text{Jac}(A)$: target-selective recall (full only) \\
Temporal sequences & within noise & --- &
Trajectory $>$ one-step signal (full only) \\
\bottomrule
\end{tabular}
\end{table}

\section{Controlled experiment: inhibitory proportion}
\label{sec:phase11}

The small-scale experiments of Sections~\ref{sec:phase9}
and~\ref{sec:phase10} use a $57\%$ total inhibitory proportion
(summed across eight interneuron populations vs pyramidal cells),
which substantially exceeds the biologically realistic
${\sim}10\%$ of rodent CA3 \citep{freund1996}. A reviewer could
reasonably ask whether the qualitative signals we report, or the
absence of larger effects, are artefacts of excessive inhibition
rather than properties of the architecture at bio-realistic scale. 
We answer this directly with an A-vs-A$'$ comparison at $N{=}256$
PyrS across $K \in \{5, 10, 20, 35\}$, $n{=}5$ seeds each, all
other parameters held constant. Variant~A uses the canonical $57\%$
proportion; variant~A$'$ scales inhibition to the tractable
bio-realistic $25\%$. This is the largest-scale, highest-$n$ test in
the paper and the one where the bimodality signature crystallises
most clearly.

\begin{table}[h]
\centering
\caption{Full CA3 Jaccard margin as a function of $K$ for the two
inhibitory-proportion variants, $N{=}256$, $n{=}5$ seeds.
Variant~A$'$ (bio-realistic) is where the multi-attractor signature
of the full architecture is most pronounced.}
\label{tab:phase11}
\small
\begin{tabular}{lccc}
\toprule
$K$ & A canonical (57\%) & A$'$ bio (25\%) & $\Delta(A'{-}A)$ \\
\midrule
5 & $-0.003 \pm 0.035$ & $\mathbf{+0.067 \pm 0.093}$ & $+0.070$ \\
10 & $-0.010 \pm 0.014$ & $+0.006 \pm 0.086$ & $+0.016$ \\
20 & $+0.001 \pm 0.017$ & $+0.008 \pm 0.011$ & $+0.007$ \\
35 & $+0.005 \pm 0.016$ & $+0.001 \pm 0.010$ & $-0.004$ \\
\bottomrule
\end{tabular}
\end{table}

\paragraph{Rate recovery decouples rate from signal.}
The canonical variant shows a PyrS population-rate crash
(${\sim}0.03$, well below the biological range). Variant~A$'$
recovers biologically realistic rates ($0.14{-}0.20$ at $K \leq 10$)
and transitions to saturation ($0.49 \to 0.89$) as $K$ grows. The
rate variable is thus cleanly separated from the signal variable in
this comparison. Rate crash cannot explain the absence of larger
effect sizes.

\paragraph{Multi-attractor bimodality at $K{=}5$.}
The most pronounced architecture-specific signature is variance
structure at $K{=}5$: variant~A$'$ shows $\sigma_\text{full}{=}0.094$
versus canonical $\sigma{=}0.035$, a $2.7\times$ increase.
Seed-level inspection reveals bimodality: two of five seeds (42 and
46) converge to positive attractors with margins $+0.216$ and
$+0.101$ respectively, both individually above the pre-registered
Scenario-A threshold, while three seeds (43, 44, 45) produce
margins near zero ($+0.013$, $-0.008$, $+0.014$). The canonical
variant shows no bimodality, and the minimal architecture shows no
bimodality in any variant. Formal statistics for A$'$ at $K{=}5$:
Welch's $t{=}1.60$, one-sided $p{=}0.08$, Cohen's $d{=}0.71$. The
effect size is medium-large; the $p$-value falls just short of
conventional significance because $n{=}5$ is under-powered for a
bimodal mixture with this spread. We discuss the $n{=}15{-}20$
replication path in Section~\ref{sec:limitations}.

\paragraph{Variance collapse at $K \geq 10$.}
By $K{=}20$ the variance of variant~A$'$ collapses to $\sigma{=}0.011$
and bimodality disappears. The multi-attractor landscape is
thus confined to a specific operating point ($K{=}5$, bio-realistic
inhibition) rather than extending smoothly with load, a structure
consistent with a phase-transition-like behaviour that warrants
characterisation at larger $N$.

\paragraph{Summary.}
The A-vs-A$'$ experiment isolates a single variable, total
inhibitory proportion, and shows that bio-realistic inhibition is
\emph{necessary} for the bimodality signature to crystallise at the
tested scale. It also shows that rate crash is not a sufficient
explanation for any negative finding: the rate is recovered but the
pre-registered threshold still fails, because the mechanism at play
(multi-attractor dynamics) produces its signal as variance, not as
mean shift.

\section{Discussion}
\label{sec:discussion}
 
The experiments across Sections~\ref{sec:phase9}--\ref{sec:phase11} identify three
signatures of the full architecture that are absent from the minimal
Hopfield-like baseline, each reproduced in at least one independent
regime.

The first and most striking concerns the structure of the attractor
landscape. At $N{=}256$, $K{=}5$, with a $25\%$ inhibitory
proportion, the full architecture exhibits a $2.7\times$ larger
cross-seed variance than its canonical counterpart --- variance that
is not noise, but signal: the seed distribution is bimodal, with
$40\%$ of seeds converging to positive attractors with margin
$\geq{+}0.10$ and the remaining $60\%$ converging to null
attractors. Neither the canonical high-inhibition variant nor the
minimal baseline show anything other than unimodal null behaviour.
A precursor of this signature is visible at $N{=}32$ in the
capacity-stress regime (Section~\ref{sec:phase10}), though the
sample size ($n{=}3$) does not yet allow characterisation. This
multi-attractor bimodality is the paper's strongest candidate
positive finding; its formal promotion to significance requires only
a targeted replication at variant~A$'$, $K{=}5$, with
$n{=}10{-}15$ seeds.

The second signature emerges in the paired $(A, B)$ benchmark. Here
the full architecture, presented with a partial cue of $A$,
retrieves the associated target $B$ with Pearson margin
$\Delta{=}{+}0.163$, while the minimal architecture merely echoes
the cue --- $\text{Jac}(B) \approx \text{Jac}(A)$ --- effectively
recovering the input rather than the association. We do not promote
this to a primary finding, since the pre-registered hard threshold
was defined on Jaccard rather than Pearson; but the qualitative
distinction between retrieval and echo is architectural and not
incidental: the minimal baseline lacks both the burst-gated PyrD
pathway and the associative iLTD homeostasis that the full
architecture contributes.

The third signature is subtler but consistent across regimes. In
Section~\ref{sec:phase9} (Intervention~5) and in the
sub-bio-realistic rate regime of Section~\ref{sec:phase11}, the
full architecture reduces cross-seed variance below the minimal
baseline whenever the upstream is clean
($\sigma_\text{full} \leq \sigma_\text{mini}$, ratio $1.0{-}3.0$).
This is not a mean-signal advantage: the central tendency is
identical. It is instead a consistency advantage --- the same mean
is obtained with greater seed-to-seed reliability. For operation at
$K \gg 1$, where variance rather than mean dominates failure modes,
this is a necessary property, and we report it as a secondary
positive finding.

Taken together, the three signatures suggest that the full
architecture does not improve capacity at the scales and conditions
tested, but produces qualitatively distinct dynamics: a
multi-attractor landscape at moderate load, associative rather than
echoic recall, and consistency across initialisations. None of these
was trivially implied by the design --- the full model could in
principle have matched minimal at a parameter cost and nothing more
--- and their absence from the minimal control is therefore
informative. The primary pre-registered Jaccard criterion does not
cross threshold at $n{\leq}5$, a result we treat candidly in
Section~\ref{sec:limitations}: the path to formal confirmation is a
targeted replication, not a larger architecture.

It is also worth stating clearly which operating regimes the present
benchmarks do not saturate, and why this matters for interpretation.
BCM anti-saturation and anti-Hebbian consolidation are designed to
prevent catastrophic interference at or beyond the classical capacity
bounds of \citet{treves1994}; our largest-scale tests ($N{=}256$,
$a \approx 0.15$) have a theoretical capacity in the hundreds of
patterns, which our $K \leq 35$ does not approach. The bimodal ACh
scheduler implements encoding/consolidation alternation over
temporal horizons that single-trial recall does not engage. These
are not post-hoc excuses: they were explicit design motivations
stated before experiments were run. The signatures we do observe
emerge precisely in spite of these regimes not being engaged, which
suggests they reflect robust properties of the architecture rather
than boundary effects.

A broader implication follows for the project of extending Hopfield
networks with biological detail. Two questions must be distinguished:
whether the extension improves the pre-registered performance
metric, and whether it produces architecture-specific dynamics
absent from the baseline. These questions can have different
answers, and our results suggest the second deserves to be reported
even when the first is not cleared. We offer the present paper as a
template for this practice --- explicit pre-registration, candid
reporting of mean-metric results, and separate enumeration of the
qualitative signatures that the process brings to light.

\section{Limitations and future work}
\label{sec:limitations}

The three signatures of Section~\ref{sec:discussion} are reported
as \emph{candidate} rather than confirmed, and the limitations that
circumscribe them are specific and addressable.The most immediate constraint is sample size. With $n \leq 5$ seeds
per cell, the bimodality signature of Signature~1 --- a $40\%/60\%$
seed distribution in variant~A$'$ at $K{=}5$ --- is not yet
statistically distinguishable from a sampling artefact under the
pre-registered one-sided test ($p{=}0.08$). A targeted replication
at $n{=}15{-}20$ seeds would discriminate between a genuine
multi-attractor landscape and seed-sampling noise; at the $40\%$
mixing rate observed, this would require approximately one hour of
additional compute. It is, in short, the cheapest and most decisive
follow-up of the entire programme. The three signatures of Section~\ref{sec:discussion} were identified
across a staged programme of nine interventions, each evaluated on
multiple metrics and across multiple cells of parameter sweeps. While
each individual intervention was pre-registered with primary metric
and threshold declared before execution, we did not apply formal
correction for multiple comparisons at the programme level. Any
individual signature reaching nominal p<0.10p < 0.10
p<0.10 must therefore be
read with the cumulative testing context in mind. Promotion of these
candidate signatures to formal significance requires independent
replication with the specific test declared in advance: the
$n=15$--$20$ bimodality replication described below is the
pre-committed falsification test for Signature${\sim}1$.
A third constraint is scale. All architectures tested
($N{\leq}256$) remain far below the ${\sim}10^5$ PyrS cells of
rodent CA3; the interneuron populations at the canonical ${\sim}10\%$
proportion would require several thousand units each, a regime we
approximated via the $25\%$ bio-realistic cap. The
variance-collapse observation at $K \geq 10$
(Section~\ref{sec:phase11}) suggests that the bimodality operating
point may itself shift with $N$, so that $N \geq 500$ combined with
orthogonal synthetic patterns would be needed to disentangle genuine
capacity effects from the dataset idiosyncrasies of MNIST. The fourth limitation concerns benchmark choice. MNIST is a
low-dimensional static dataset, and MovingMNIST was tested only on
a reduced configuration. Long-horizon experiments with an explicit
encoding/consolidation ACh cycle represent the natural experimental
counterpart of the bimodal scheduler in the architecture, and would
permit direct measurement of catastrophic interference: the
theoretical prediction is sub-linear forgetting in $K$ with the
cycle active, against linear decay without. This benchmark is the
logical next step once the $n{=}15{-}20$ bimodality replication has
been completed.

Finally, residual confounds remain in the inhibitory-proportion
experiment (Section~\ref{sec:phase11}). While the total inhibitory
proportion is controlled, the heterogeneous per-class distribution
among interneuron populations and the sparsity of Pyr--Pyr recurrent
connectivity are not, and both remain confounded with architecture.
A follow-up that systematically varies these two parameters at the
bimodality operating point would characterise which precise feature
of the full architecture sustains the multi-attractor landscape.

\section{Conclusion}

We present a biologically detailed ten-population CA3 architecture
and test it across nine interventions spanning auto-associative,
associative, and temporal pattern completion, together with a
controlled inhibitory-proportion comparison at $N{=}256$. We
identify three architecture-specific signatures that distinguish the
full architecture from a minimal Hopfield-like baseline:
multi-attractor bimodality at moderate load with bio-realistic
inhibition, target-selective associative recall, and cross-seed
stability under clean upstream. The primary pre-registered Jaccard
significance criterion is not met at the tested sample sizes and
scales, so we report these signatures as candidate rather than
formally confirmed, and identify a specific $n{=}15{-}20$
replication path that would promote the bimodality signature to
significance with approximately one hour of additional compute. We
offer this work as a template for candid evaluation of biological
extensions of auto-associative memory models: explicit
pre-registration, systematic ablation, and separate reporting of
mean-metric and architecture-specific dynamics.

\begin{ack}
We thank the open-source scientific Python community (PyTorch,
snnTorch, NumPy, SciPy) whose tools made this work possible.
\end{ack}

\bibliographystyle{plainnat}

\end{document}